\documentclass[11pt,a4paper,dvipsnames]{article}
\usepackage[hyperref]{emnlp2020}
\usepackage{times}
\usepackage{latexsym}

\usepackage{microtype}

\aclfinalcopy 

\usepackage[fleqn]{amsmath}
\usepackage{amsfonts}
\usepackage{booktabs}
\usepackage{multirow}
\usepackage{substitutefont}
\usepackage{textgreek}
\usepackage{xcolor}

\usepackage{tikz}
\usetikzlibrary{bayesnet}

\substitutefont{LGR}{ptm}{cmr}

\newcommand{\red}[1]{\textcolor{BrickRed}{#1}}
\newcommand{\blue}[1]{\textcolor{MidnightBlue}{#1}}
\newcommand{\ent}{\texttt{\textless ENT\textgreater } }

\newcommand{\fonttable}{\fontsize{10}{12}\selectfont}

\title{Learning Informative Representations of Biomedical Relations with Latent Variable Models}

\author{Harshil Shah \\
  University College London \\
  BenevolentAI\thanks{\; Work completed during internship at BenevolentAI.} \\
  \texttt{h.shah@cs.ucl.ac.uk} \\\And
  Julien Fauqueur \\
  BenevolentAI \\
  \texttt{julien@benevolent.ai}}

\date{}

\begin{document}
\maketitle
\begin{abstract}
    Extracting biomedical relations from large corpora of scientific documents is a challenging natural language processing task. Existing approaches usually focus on identifying a relation either in a single sentence (mention-level) or across an entire corpus (pair-level). In both cases, recent methods have achieved strong results by learning a point estimate to represent the relation; this is then used as the input to a relation classifier. However, the relation expressed in text between a pair of biomedical entities is often more complex than can be captured by a point estimate. To address this issue, we propose a latent variable model with an arbitrarily flexible distribution to represent the relation between an entity pair. Additionally, our model provides a unified architecture for both mention-level and pair-level relation extraction. We demonstrate that our model achieves results competitive with strong baselines for both tasks while having fewer parameters and being significantly faster to train. We make our code publicly available.
\end{abstract}

\section{Introduction} \label{sec:intro}

The vast amounts of scientific literature can provide a significant source of information for biomedical research. Using this literature to identify relations between entities is an important task in various applications \citep{VFGMNTKF2012,SMH2013,BPQRF2015,KRAPSRTILNVCRLDOLV2017}.

Existing approaches to biomedical relation extraction usually fall into one of two categories. Mention-level extraction aims to classify the relation between a pair of entities within a short span of text (usually a sentence). In contrast, pair-level extraction aims to classify the relation between a pair of entities across an entire paragraph, document or corpus.

For both mention-level and pair-level relation extraction, recent work has been focused on representation learning. This is considered to be one of the major steps towards making progress in artificial intelligence \citep{BCV2013}. Representations of relations which understand their context are particularly important in biomedical research, where identifying fruitful targets is crucial due to the high costs of experimentation. Learning such representations is likely to require large amounts of unsupervised data due to the scarcity of labelled data in this domain.

Recent mention-level methods have been based on using large unsupervised models with Transformer networks \citep{VSPUJGKP2017} to learn representations of sentences containing pairs of entities. These representations are then used as the inputs to much smaller models, which perform supervised relation classification \citep{LYKKKSK2019,BLC2019}.

Recent pair-level methods have been based on encoding each mention of a pair of entities, and designing a mechanism to pool these encodings (across a paragraph, document, or corpus) into a single representation. This representation is then used to classify the relation between the entity pair \citep{BRAN2018,JWP2019}.

However, representation learning methods for both mention-level and pair-level extraction typically use a point estimate for each representation. As a result, they may struggle to capture the nature of the true, potentially complex relations between each pair of entities. For example, Figure \ref{fig:intro:examples} shows sentences for two entity pairs which demonstrate that relation statements can be very different, typically depending on biological circumstances (\textit{e.g.} anatomical location, experimental details, presence of a disease, \textit{etc}). Such nuanced relations can be difficult to capture with a single point estimate.

We hypothesise that there is a true underlying relation for each entity pair, and that this relation can be multimodal (because of the aforementioned complexities). The sentences containing each pair are textual observations of these underlying relations.

We therefore propose a probabilistic model which uses a continuous latent variable to represent the true relation between each entity pair. The distribution of a sentence containing that pair is then conditioned on this latent variable. In order to be able to model the complex relations between each entity pair, we use an infinite mixture distribution for the latent representation.

Our model provides a unified architecture for learning representations of relations between entity pairs both at mention and pair level. We show that (an approximation to) the posterior distribution of the latent variable can be used for mention-level relation classification. We also demonstrate that the prior distribution from the same model can be used for pair-level classification. On both tasks, we achieve results competitive with strong baselines with a model which has fewer parameters and is significantly faster to train.

The code is released at \url{ https://github.com/BenevolentAI/RELVM}

\begin{figure}[t]
    \fonttable
    \centering
    \begin{tabular}{p{0.92\linewidth}}
        \toprule
        Protein \textbf{\red{Akt}} and protein \textbf{\blue{GSK3\textbeta}}: \\
        \midrule
        ``\ldots \ \textbf{\red{Akt}} negatively regulates \textbf{\blue{GSK3\textbeta}} activity\ldots'' \\[2pt]
        ``\ldots \ \textbf{\red{Akt}} phosphorylates \textbf{\blue{GSK3\textbeta}}\ldots'' \\
        \bottomrule
    \end{tabular}
    
    \vspace{12pt}
    
    \begin{tabular}{p{0.92\linewidth}}
        \toprule
        Protein \textbf{\red{EAAT2}} and disease \textbf{\blue{ALS}}: \\
        \midrule
        ``\textbf{\red{EAAT2}}/C1-4 were found to be equally expressed in \textbf{\blue{ALS}} patients and controls.'' \\[2pt]
        ``\textbf{\red{EAAT2}} protein is significantly reduced in \textbf{\blue{ALS}} in the motor cortex and spinal cord.'' \\
        \bottomrule
    \end{tabular}
    \caption{Two sets of sentences demonstrating the potentially complex nature of the relation between a pair of entities.}
    \label{fig:intro:examples}
    \vskip -8pt
\end{figure}

\section{Model} \label{sec:model}

In this section, we introduce our unified architecture for both mention-level and pair-level relation extraction. Throughout, we use the following notation: \begin{itemize}
    \item $c$ represents a `context', \textit{i.e.} a sentence (or sequence of tokens) containing a pair of entities. $c$ has tokens $c_{1}, \hdots, c_{T}$.
    \begin{itemize}
        \item $c_{t_{x}}$ and $c_{t_{y}}$ are the tokens representing the two entities. We replace the actual tokens denoting the two entities with generic \ent tokens. Therefore, a context is given by: \begin{align*}
            c = \ & c_{1}, \hdots, c_{t_{x} - 1}, \ent, c_{t_{x} + 1}, \hdots, \\
            & c_{t_{y} - 1}, \ent, c_{t_{y} + 1}, \hdots, c_{T}
        \end{align*}
    \end{itemize}
    
    \item $x$ and $y$ are the input representations of the two entities. 
    \begin{itemize}
        \item For pair-level classification, $x$ and $y$ will be unique identifiers for the two entities.
        \item For mention-level classification, $x$ and $y$ will be the types of the two entities, \textit{e.g.} \texttt{GENE} and \texttt{DISEASE}. This is done in order to allow for fair comparisons with previous methods, which use the entity types for mention-level classification (see Section \ref{sec:experiments:supervised} for further details).
        \item $x$ and $y$ always refer to the first and second entities in $c$ respectively. 
    \end{itemize}
    \item $\mathbf{e}(c_{t})$ is the embedding of token $c_{t}$. $\mathbf{e}(x)$ and $\mathbf{e}(y)$ are the embeddings of the entities $x$ and $y$.
    \item $r$ represents the relation label.
\end{itemize}

\paragraph{Approach}

Large corpora of labelled relation statements are often scarce, whereas unlabelled sentences are usually plentiful. In order to leverage these unlabelled sentences, we first train an unsupervised model to learn representations of entity pairs and the contexts in which they occur. We then train much smaller models to classify relations using the representations from the unsupervised model.

\subsection{Representation learning model} \label{sec:model:unsup}

When training the unsupervised representation learning model, we assume access to a corpus of sentences in which entities have been tagged but there are no relation labels. We train the representation model to maximise the conditional log-likelihood $\log p(c|x,y)$. $\theta$ will refer to the set of parameters of the representation model which we wish to optimise. A graph of the representation model is shown in Figure \ref{fig:model:unsup:graph} and a more detailed explanation is given below.

\begin{figure}[t]
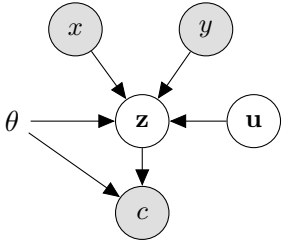

    \centering
    \tikz{%
        \node[obs] (x) {$x$};
        \node[obs, right=of x] (y) {$y$};
        \node[latent, below=0.5 of x, xshift=25pt] (z) {$\mathbf{z}$};
        \node[latent, right=0.75 of z] (u) {$\mathbf{u}$};
        \node[obs, below=0.5 of z] (c) {$c$};
        \node[const, left=1.25 of z] (theta) {$\theta$};
        \edge {x} {z};
        \edge {y} {z};
        \edge {u} {z};
        \edge {z} {c};
        \edge[shorten <= 4pt] {theta} {z};
        \edge[shorten <= 4pt] {theta} {c};
    }
    \caption{A graph depicting our unsupervised representation learning model. Clear nodes denote latent variables and shaded nodes denote observed variables. Representations from this model are used for both mention-level and pair-level relation classification.}
    \label{fig:model:unsup:graph}
\end{figure}

There are many ways to express the same relation between a given pair of entities. For example, the sentences ``\textit{John is Mary's brother}'' and ``\textit{Mary is John's sister}'' express the same relation in different ways. In order to capture this phenomenon, we introduce a latent variable, $\mathbf{z}$, to represent the true underlying relation. This will be the representation used for mention-level and pair-level relation classification. The conditional distribution is parametrised as: \begin{align}
    p(c|x,y) = \int_{\mathbf{z}} p_{\theta}(\mathbf{z}|x,y) p_{\theta}(c|\mathbf{z})
\end{align}
Intuitively, $p_{\theta}(\mathbf{z}|x,y)$ captures the true underlying relation between the two entities $x$ and $y$, and $p_{\theta}(c|\mathbf{z})$ captures the variation in the multiple possible ways of expressing that relation.

For computational simplicity, we could choose $p_{\theta}(\mathbf{z}|x,y)$ to be Gaussian. However in reality, the true relation between a pair of entities is probably more complex than can be modelled well with a unimodal distribution. We therefore introduce another latent variable $\mathbf{u}$ such that: \begin{align}
    p_{\theta}(\mathbf{z}|x,y) = \int_{\mathbf{u}} p(\mathbf{u}) p_{\theta}(\mathbf{z}|x,y,\mathbf{u})
\end{align}
For $p(\mathbf{u})$, we use a standard Gaussian distribution, $\mathcal{N}(\mathbf{0}, \mathbf{I})$. For $p_{\theta}(\mathbf{z}|x,y,\mathbf{u})$, we again use a Gaussian distribution whose mean and variance are a function of $x$, $y$ and $\mathbf{u}$. We concatenate together $\mathbf{e}(x)$, $\mathbf{e}(y)$, $\mathbf{e}(x) \odot \mathbf{e}(y)$ and $\mathbf{u}$, and pass the resulting vector into a feedforward network to output the mean and variance of $p_{\theta}(\mathbf{z}|x,y,\mathbf{u})$ ($\odot$ denotes element-wise multiplication). Using a nonlinear network allows the marginal distribution $p_{\theta}(\mathbf{z}|x,y)$ to be an infinite mixture distribution \citep{MF2018}. The objective becomes: \begin{align}
    \log p(c|x,y) = \log \int_{\mathbf{u},\mathbf{z}} p(\mathbf{u}) p_{\theta}(\mathbf{z}|x,y,\mathbf{u}) p_{\theta}(c|\mathbf{z}) \label{eq:model:unsup:obj}
\end{align}
We parametrise $p_{\theta}(c|\mathbf{z})$ with an LSTM, due to its strong performance in language modelling \citep{Graves_RNN_2013,BVVDJB2016,SOTALM_Melis_2018}. The conditional probabilities for $t = 1, \hdots, T$ are: \begin{align}
    p_{\theta}(c_{t}=v|c_{1:t-1},\mathbf{z}) \propto \exp((\mathbf{W}\mathbf{h}^{p}_{t}) \cdot \mathbf{e}(v)) \label{eq:model:unsup:context}
\end{align}
where $\mathbf{W}$ is a learnable parameter of the model, and $\mathbf{h}_{t}^{p}$ is computed as: \begin{align}
    \mathbf{h}^{p}_{t} = \textrm{LSTM}(\mathbf{z}, \mathbf{h}^{p}_{t-1}, \mathbf{e}(c_{t-1})) \label{eq:model:unsup:context:rnn}
\end{align}
Complete hyperparameter details are provided in Section \ref{sec:experiments:unsupervised}.

\subsubsection{Training} \label{sec:model:unsup:training}

Because of the nonlinear functions involved in $p_{\theta}(\mathbf{z}|x,y,\mathbf{u})$ and $p_{\theta}(c|\mathbf{z})$, the integral in Equation (\ref{eq:model:unsup:obj}) is intractable. We therefore perform approximate maximum likelihood estimation using stochastic gradient variational Bayes (SGVB) \citep{KW2014,RMW2014}.

To do this, we parametrise a Gaussian inference distribution $q_{\phi}(\mathbf{u}|x,y,c)$ (referred to as $q_{\phi}(\mathbf{u})$ henceforth, for brevity) with trainable parameters $\phi$. This allows us to maximise the following lower bound on the log-likelihood: \begin{align}
    \log p(c|x,y) \! & \geq \mathbb{E}_{q_{\phi}(\mathbf{u}) p_{\theta}(\mathbf{z}|x,y,\mathbf{u})} \! \! \left[ \log \frac{p(\mathbf{u}) p_{\theta}(c|\mathbf{z})}{q_{\phi}(\mathbf{u})} \right] \nonumber \\
    & \equiv \mathcal{L}_{\theta,\phi}(c,x,y) \label{eq:model:unsup:obj:lb}
\end{align}
This bound can be approximated using Monte Carlo integration. It is optimised with respect to $\theta$ and $\phi$ jointly.

To parametrise $q_{\phi}(\mathbf{u})$, we use a bidirectional LSTM to encode the context. This is due to its ability to capture useful sentence level information into a low-dimensional vector \citep{zhou-etal-2016-text,peters-etal-2018-deep}. It is computed as: \begin{align}
    & \overrightarrow{\mathbf{h}^{q}_{t}} = \mathrm{LSTM}(\mathbf{e}(c_{t}), \overrightarrow{h^{q}_{t-1}}) \\
    & \overleftarrow{\mathbf{h}^{q}_{t}} = \mathrm{LSTM}(\mathbf{e}(c_{t}), \overleftarrow{h^{q}_{t+1}}) \\
    & \mathbf{h}^{q} = [\overrightarrow{\mathbf{h}^{q}_{T}}; \overleftarrow{\mathbf{h}^{q}_{1}}]
\end{align} 
We concatenate $\mathbf{h}^{q}$ to $\mathbf{e}(x)$, $\mathbf{e}(y)$ and $\mathbf{e}(x) \odot \mathbf{e}(y)$ and pass the resulting vector into a feedforward network to output the mean and variance of $q_{\phi}(\mathbf{u})$.

\subsection{Mention-level classification} \label{sec:model:sup}

In this section, we assume that the unsupervised representation model from Section \ref{sec:model:unsup} has been trained with $x$ and $y$ being the types of the two entities. The representations $\mathbf{z}$ can now be used as the inputs to a supervised mention-level relation classification model. 

For mention-level classification, we assume access to a corpus of sentences in which entities have been tagged and there are labels classifying the type of relation between the entity pair in each sentence. We train the mention-level classification model to maximise $p(r|x,y,c)$. $\lambda$ will refer to the set of parameters of the mention-level classification model which we wish to optimise.

The representation $\mathbf{z}$ of the entity pair and context would ideally be distributed according to the posterior $p(\mathbf{z}|x,y,c)$ from the representation model. We would then optimise the parameters $\lambda$ using the following objective: \begin{align}
    p(r|x,y,c) = \int_{\mathbf{z}} p(\mathbf{z}|x,y,c) p_{\lambda}(r|\mathbf{z})
\end{align}
However: \begin{align}
    p(\mathbf{z}|x,y,c) &= \frac{p_{\theta}(\mathbf{z},c|x,y)}{p(c|x,y)} \\
    &= \frac{p_{\theta}(\mathbf{z},c|x,y)}{\int_{\mathbf{u}, \mathbf{z}} p(\mathbf{u}) p_{\theta}(\mathbf{z}|x,y,\mathbf{u}) p_{\theta}(c|\mathbf{z})}
\end{align}
As mentioned in Section \ref{sec:model:unsup:training}, the integral in the denominator is intractable. Instead, the following approximation to the posterior can be used: \begin{align}
    p(\mathbf{z}|x,y,c) \simeq \int_{\mathbf{u}} q_{\phi}(\mathbf{u}) p_{\theta}(\mathbf{z}|x,y,\mathbf{u})
\end{align}
This is an approximation to the posterior because maximising the objective in Equation (\ref{eq:model:unsup:obj:lb}) is equivalent to minimising the KL divergence from $q_{\phi}(\mathbf{u}) p_{\theta}(\mathbf{z}|x,y,\mathbf{u})$ to $p(\mathbf{z},\mathbf{u}|x,y,c)$ \citep{KW2014}: \begin{align}
    & \mathcal{L}_{\theta,\phi}(c,x,y) = \log p(c|x,y) - \nonumber \\
    & \quad D_{\mathrm{KL}}[q_{\phi}(\mathbf{u}) p_{\theta}(\mathbf{z}|x,y,\mathbf{u}) || p(\mathbf{z},\mathbf{u}|x,y,c)]
\end{align}
Using this approximation, the mention-level classification objective becomes: \begin{align}
    p(r|x,y,c) \! & \simeq \! \! \int_{\mathbf{u}, \mathbf{z}} q_{\phi}(\mathbf{u}) p_{\theta}(\mathbf{z}|x,y,\mathbf{u}) p_{\lambda}(r|\mathbf{z}) \\
    &= \mathbb{E}_{q_{\phi}(\mathbf{u}) p_{\theta}(\mathbf{z}|x,y,\mathbf{u})} [p_{\lambda}(r|\mathbf{z})] \label{eq:model:sup:obj}
\end{align}
Empirically, however, we find that the model trains much more easily using the following objective: \begin{align}
    \mathbb{E}_{q_{\phi}(\mathbf{u}) p_{\theta}(\mathbf{z}|x,y,\mathbf{u})} [\log p_{\lambda}(r|\mathbf{z})] \! \equiv \! \mathcal{L}_{\lambda}(r,c,x,y) \label{eq:model:sup:obj:lb}
\end{align}
This is due, particularly at the start of training, to the values of $p_{\lambda}(r|\mathbf{z})$ being very small. Note that, due to Jensen's inequality, the objective in Equation (\ref{eq:model:sup:obj:lb}) is in fact a lower bound on the $\log$ of the objective in Equation (\ref{eq:model:sup:obj}): \begin{align}
    \mathcal{L}_{\lambda}(r,c,x,y) \! \leq \! \log \mathbb{E}_{q_{\phi}(\mathbf{u}) p_{\theta}(\mathbf{z}|x,y,\mathbf{u})} [p_{\lambda}(r|\mathbf{z})]
\end{align}
To parametrise $p_{\lambda}(r|\mathbf{z})$, we use a shallow feedforward network with a softmax function at the output. Complete hyperparameter details are provided in Section \ref{sec:experiments:supervised}.

\subsection{Pair-level classification}  \label{sec:model:pair}

In this section, we assume that the unsupervised representation model from Section \ref{sec:model:unsup} has been trained with $x$ and $y$ being unique identifiers for the two entities. The representations $\mathbf{z}$ can now be used as the inputs to a supervised pair-level relation classification model. 

For pair-level classification, we assume access to a dataset with pairs of entity identifiers, and labels classifying the type of relation between each pair. Instead of learning $p(r|x,y,c)$ as in mention-level classification, we now learn $p(r|x,y)$.

Intuitively, for pair-level classification, we wish to classify the relation between a pair of entities based on everything that the unsupervised model has learned about those entities (through the sentences containing them). This is unlike mention-level classification, where we classify the relation described in a specific sentence.

For pair-level classification, we follow a very similar approach to that described in Section \ref{sec:model:sup} for mention-level classification. However we no longer base the input representation on the posterior distribution from the unsupervised model, $p(\mathbf{z}|x,y,c)$. Instead, the representation used will be distributed according to: \begin{align}
    p_{\theta}(\mathbf{z}|x,y) = \int_{\mathbf{u}} p(\mathbf{u}) p_{\theta}(\mathbf{z}|x,y,\mathbf{u})
\end{align}
Intuitively, this is the natural distribution to use, because we are interested in the relation between the entities $x$ and $y$, without a specific context to condition on.

We denote $\psi$ as the parameters of the pair-level supervised model. Then, following the same reasoning as Section \ref{sec:model:sup}, the objective for the pair-level supervised model is: \begin{align}
    \mathbb{E}_{p(\mathbf{u}) p_{\theta}(\mathbf{z}|x,y,\mathbf{u})} [\log p_{\psi}(r|\mathbf{z})] \equiv \mathcal{L}_{\psi}(r,x,y) \label{eq:model:pair:obj:lb}
\end{align}
To parametrise $p_{\psi}(r|\mathbf{z})$, we use a shallow feedforward network with a softmax function at the output. Complete hyperparameter details are provided in Section \ref{sec:experiments:pair}.

\section{Related work} \label{sec:related}

Mention-level relation extraction is typically performed using supervised learning. In the general domain, \citet{tacred2017} combine an LSTM with a position-aware attention mechanism to perform multiclass relation extraction. \citet{SFLK2019} fine-tune the BERT \citep{DCLT2019} architecture to relation extraction tasks by enforcing similarity between representations of sentences containing the same pair of entities across a corpus. \citep{Distill_Zhang_2020} construct a teacher model to generate soft labels which guide the optimisation of a student network via knowledge distillation. In the biomedical and scientific domains, BioBERT \citep{LYKKKSK2019} and SciBERT \citep{BLC2019} train the BERT architecture on domain-specific corpora, achieving state of the art results on mention-level relation extraction tasks. \citet{zhang-hybrid-2018} combine an RNN over the sentence's words and a CNN over its dependency graph to classify drug-drug and protein-protein interactions.

Pair-level relation extraction usually relies on distant supervision \citep{mintz-etal-2009-distant}. In the general domain, \citet{multir2011} develop a latent variable model to perform multi-instance learning while handling overlapping relations. \citet{LSLLS2016} use an attention mechanism to pool the representations of sentences containing a given pair into a single representation, which is then used as the input to a classifier. \citet{QuirkPoon16} capture relations across sentences by linking dependency graphs between sentences. Other pair-level methods build representations using unsupervised models. \citet{ijcai2019relationvectors} use a latent variable model to learn a point-estimate representation from the unigram distribution of tokens co-occurring in sentences with the given pair. \citet{pair2vec2019} learn representations of pairs of entities by maximising their pointwise mutual information (PMI) with the contexts that the entities appear in. In the biomedical domain, \citet{BRAN2018} build a paragraph-level representation using a modified Transformer network, and aggregate over mentions using a softmax function. \citet{knowledge-graph-cascade-2019} combine knowledge embeddings and graph embeddings using a cascade learning framework to predict links in biochemical networks. \citet{ebc-altman-2015} use a distributional semantics approach to cluster together drug-gene pairs which are related in similar ways.

Contrary to our work, there does not appear to be prior research performing both mention-level and pair-level relation extraction with a unified model.

\section{Experiments} \label{sec:experiments}

\subsection{Representation learning model} \label{sec:experiments:unsupervised}

We train the unsupervised representation model described in Section \ref{sec:model:unsup} using sentences from PubMed abstracts, PubMed Central (PMC) open-access full-text articles, and licensed full-text articles from Wiley and Springer. We take sentences with a maximum length of 140 tokens and tag the entities with their type using a dictionary-based method. Entities are linked to unique identifiers by first disambiguating entity types using a bidirectional LSTM sentence classifier, followed by type-specific term lookups. Note that if a sentence contains three or more entities, it is repeated in order to account for each possible pair of entities.

\subsubsection{Architectures and training}

To parametrise $p_{\theta}(\mathbf{z}|x,y,\mathbf{u})$, we use a 2-layer feedforward network with the ReLU nonlinearity. To parametrise $p_{\theta}(c|\mathbf{z})$, we use a 1-layer LSTM. To parametrise $q_{\phi}(\mathbf{u})$, we use a 1-layer bidirectional LSTM, the output of which is passed to a 2-layer feedforward network with the ReLU nonlinearity.

In order to evaluate the effect of the number of parameters on performance, we train four different versions of our representation learning model: \{\textsc{x-small}, \textsc{small}, \textsc{medium}, \textsc{large}\}. These correspond to respective hidden state sizes of \{128, 256, 512, 1024\} in the networks. For all of the models, both $\mathbf{u}$ and $\mathbf{z}$, as well as all embeddings, are 300-dimensional.

We train the unsupervised representation models using a single sample approximation of the objective in Equation (\ref{eq:model:unsup:obj:lb}). We train for 400,000 iterations, using a minibatch size of 192 and optimising the parameters using Adam \citep{KB2015} with a learning rate of 0.0001.

\subsubsection{Optimisation challenges}

The unsupervised objective in Equation (\ref{eq:model:unsup:obj:lb}) can be expressed as: \begin{align}
    \mathcal{L}_{\theta, \phi}(c,x,y) &= \mathbb{E}_{q_{\phi}(\mathbf{u}) p_{\theta}(\mathbf{z}|x,y,\mathbf{u})} [\log p_{\theta}(c|\mathbf{z})] \nonumber \\
    & \quad - D_{\mathrm{KL}}[q_{\phi}(\mathbf{u}) || p(\mathbf{u})] \label{eq:model:unsup:obj:lb:alt}
\end{align}
When training latent variable models with autoregressive observation distributions (such as that in Equation (\ref{eq:model:unsup:context})), this objective can induce local optima where $q_{\phi}(\mathbf{u})$ = $p(\mathbf{u})$. This results in the KL divergence term in Equation (\ref{eq:model:unsup:obj:lb:alt}) collapsing to 0, meaning the model ignores the latent variable altogether. To avoid such local optima, we use the following two methods \citep{BVVDJB2016}:

\paragraph{KL annealing}

We multiply the KL divergence term by a constant weight which is linearly annealed from 0 to 1 over the first 10,000 iterations of training. This helps the model to escape local optima where $D_{\mathrm{KL}}[q_{\phi}(\mathbf{u}) || p(\mathbf{u})] = 0$ early in training.

\paragraph{Token dropout}

In Equation (\ref{eq:model:unsup:context:rnn}), we randomly drop the token embedding being passed to the next LSTM hidden state. We use a dropout rate of 50\%. This encourages the LSTM to rely more on the representation $\mathbf{z}$ than the previous tokens when modelling the context.

\subsubsection{Computational costs}

We show the computational costs of our unsupervised representation models in Table \ref{tab:experiments:unsupervised:compcosts}. We compare against BioBERT \citep{LYKKKSK2019}, a language model with state-of-the-art performance on relation extraction.

All versions of our model have significantly fewer parameters than BioBERT. In terms of `GPU days'\footnotemark, training BioBERT is approximately 25 to 40 times slower than training our model. In addition, inference is an order of magnitude faster with our model compared to BioBERT.

\footnotetext{GPU days = No. of GPUs $\times$ training time (in days).}

\begin{table*}[t]
    \centering
    \begin{tabular}{ll @{\hskip 24pt} lc @{\hskip 24pt} lc}
        \toprule
         & & \multicolumn{2}{c}{\textsc{Training}} & \multicolumn{2}{c}{\textsc{Inference}} \\
        \textsc{Model} & \textsc{Params} & \textsc{Hardware} & \textsc{Time} & \textsc{Hardware} & \textsc{Time} \\
        \midrule
        BioBERT                 & 110M & 8 x V100 GPUs & 10 days & 1 x V100 GPU & 0.0087s/sent. \\
        Ours (\textsc{x-small}) &   2M & 1 x V100 GPU  &  2 days & 1 x V100 GPU & 0.0004s/sent. \\
        Ours (\textsc{small})   &   4M & 1 x V100 GPU  &  2 days & 1 x V100 GPU & 0.0004s/sent. \\
        Ours (\textsc{medium})  &  10M & 1 x V100 GPU  &  3 days & 1 x V100 GPU & 0.0005s/sent. \\
        Ours (\textsc{large})   &  30M & 1 x V100 GPU  &  3 days & 1 x V100 GPU & 0.0007s/sent. \\
        \bottomrule
    \end{tabular}
    \caption{The computational costs of each of the unsupervised representation models we train. The inference time for each model is computed on a V100 GPU.}
    \label{tab:experiments:unsupervised:compcosts}
\end{table*}

\subsection{Mention-level classification} \label{sec:experiments:supervised}

After training the unsupervised representation model (using the entity types for $x$ and $y$), we use it to perform supervised mention-level relation classification, as described in Section \ref{sec:model:sup}. We use the EU-ADR \citep{VFGMNTKF2012} and GAD \citep{BPQRF2015} datasets. In both datasets, each sentence contains a gene and disease. The task is to classify whether the given sentence either does or does not exhibit a relation between the gene and the disease. Examples from both datasets are shown in Table \ref{tab:experiments:supervised:data:examples} and dataset statistics are shown in Table \ref{tab:experiments:supervised:data:statistics}. As per previous work, we report the performance using 10-fold cross validation on each dataset \citep{LYKKKSK2019}.

\begin{table*}[t]
    \centering
    \begin{tabular}{lll p{8.5cm} l}
        \toprule
        \textsc{Dataset} & $x$ & $y$ & $c$ & $r$ \\
        \midrule
        \multirow{4}{*}{EU-ADR} & \texttt{GENE} & \texttt{DISEASE} & Based on \ent analyses, 41 \ent patients and 12 healthy controls were studied. & 0 \\
         & \texttt{DISEASE} & \texttt{GENE} & \ent is associated with decreased expression of mucosal \ent. & 1 \\
        \midrule
        \multirow{4}{*}{GAD} & \texttt{GENE} & \texttt{DISEASE} & A broad protective effect of \ent S180L against \ent per se is not discernible. & 0 \\
         & \texttt{GENE} & \texttt{DISEASE} & The \ent polymorphism Tyr402His appears indicative of \ent pathogenesis. & 1 \\
        \bottomrule
    \end{tabular}
    \caption{Positive ($r = 1$) and negative ($r = 0$) examples from the EU-ADR and GAD datasets.}
    \label{tab:experiments:supervised:data:examples}
    \vskip -8pt
\end{table*}

\begin{table}[t]
    \centering
    \begin{tabular}{lcc}
        \toprule
        \textsc{Dataset} & EU-ADR & GAD \\
        \midrule
        \# relations & 355 & 5330 \\
        \bottomrule
    \end{tabular}
    \caption{Number of relations for the EU-ADR and GAD datasets.}
    \label{tab:experiments:supervised:data:statistics}
    \vskip -8pt
\end{table}

We compare our results with those of BioBERT as reported by \citet{LYKKKSK2019}. For a fair comparison, we use the same classifier architecture. This is a single layer network with a softmax nonlinearity. As well as training the parameters $\lambda$ of the classifier, we also fine tune the parameters $\theta$ and $\phi$ of the representation model. Again, this is done to allow for a fair comparison with BioBERT (which follows the same procedure).

We approximate the objective in Equation (\ref{eq:model:sup:obj:lb}) using 4 samples during training. We use a minibatch size of 8 and update the parameters using Adam with a learning rate of 0.00001. We train on EU-ADR for 200 iterations and on GAD for 3,000 iterations.

Note that the representations for BioBERT are 768-dimensional. This is in contrast to ours which are 300-dimensional.

\begin{table*}[t]
    \centering
    \begin{tabular}{lcccccc}
        \toprule
        \multirow{2}{2.2cm}{\textsc{Model}} & \multicolumn{3}{c}{EU-ADR} & \multicolumn{3}{c}{GAD} \\
         & P & R & F & P & R & F \\
        \midrule
        BioBERT & 80.92 & 90.81 & 84.83 & \textbf{75.95} & 88.08 & \textbf{81.52} \\
        Ours (\textsc{x-small}) & 79.62 & 98.08 & 87.71 & 67.83 & 90.76 & 77.45 \\
        Ours (\textsc{small})   & 80.35 & 98.09 & 88.14 & 68.31 & 91.75 & 78.16 \\
        Ours (\textsc{medium})  & 80.72 & 98.46 & 88.59 & 69.68 & 91.82 & 78.72 \\
        Ours (\textsc{large})   & \textbf{82.34} & \textbf{98.85} & \textbf{89.67} & 72.26 & \textbf{92.00} & 80.79 \\
        \bottomrule
    \end{tabular}
    \caption{Results using 10-fold cross validation on the EU-ADR and GAD classification tasks. We report the mean precision (P), recall (R) and F1-score (F) over the 10 folds. For all metrics, higher is better.}
    \label{tab:experiments:supervised:results}
    \vskip -0.1in
\end{table*}

\subsubsection{Results}

We perform 10-fold cross validation, and report the mean precision, recall and F1-score in Table \ref{tab:experiments:supervised:results}. On EU-ADR, all versions of our model outperform BioBERT, with our \textsc{Large} model achieving a significantly higher F1-score. On this task, all versions of our model have significantly higher recall than BioBERT, with the precision being similar. On GAD, BioBERT slightly outperforms our \textsc{Large} model, thanks to its higher precision. In addition, we find that, on both tasks, the performance monotonically increases with the size of the unsupervised representation model.

These results show that it is possible to achieve results competitive with the state-of-the-art while making significant efficiency gains, both in terms of memory and time.

\subsection{Pair-level classification} \label{sec:experiments:pair}

In this section, we use the \textsc{large} representation model from Section \ref{sec:experiments:unsupervised}, trained using the unique entity identifiers for $x$ and $y$. We fix the parameters of the unsupervised representation model and use it to perform supervised pair-level classification, as described in Section \ref{sec:model:pair}. 

We construct a multiclass classification dataset by combining multiple third-party biomedical datasets. These datasets only provide pairs of entities which are related. Therefore, if an entity pair does not appear in any of the datasets, they are assumed to be unrelated and given the label \texttt{NO-RELATION}. If two entities are related, the label is given by the concatenation of the two entity types. This is therefore a multiclass classification problem, with the set of possible classes being \{\texttt{NO-RELATION}, \texttt{DISEASE-GENE}, \texttt{GENE-GENE}, \texttt{CHEMICAL-GENE}, \texttt{CHEMICAL-DISEASE}\}. Note that we only include entity pairs that occur in at least one sentence in the dataset used to train the representation learning model.

We randomly split the related entity pairs into training, validation and test sets. The set of entity pairs with label \texttt{NO-RELATION} is extremely large. We randomly assign a proportion of these to the validation and test sets. During training, we randomly sample a proportion of each minibatch from the remaining unrelated entity pairs. The dataset statistics are shown in Table \ref{tab:experiments:pair:data:statistics}.

\begin{table}[t]
    \centering
    \begin{tabular}{lc}
        \toprule
        \textsc{Dataset} & \textsc{Pair-level} \\
        \midrule
        Train (excl. \texttt{NO-RELATION}) & 263,112 \\
        Validation & 691,627 \\
        Test & 692,534 \\
        \bottomrule
    \end{tabular}
    \caption{Total counts across all relation types for the pair-level classification dataset. The training set excludes \texttt{NO-RELATION} types, as these are sampled during training.}
    \label{tab:experiments:pair:data:statistics}
\end{table}

For the pair-level classifier, we train a 2-layer model which has a 300-dimensional hidden layer with a skip connection. We approximate the objective in Equation (\ref{eq:model:pair:obj:lb}) using 4 samples during training. We train for 100,000 iterations, using a minibatch size of 512 (of which 448 are sampled from the \texttt{NO-RELATION} set). We optimise the parameters using Adam with a learning rate of 0.0001. When making predictions on unseen data points, we only predict a label other than \texttt{NO-RELATION} if the predicted probability is higher than a threshold. This threshold is tuned to maximise the F1-score on the validation set.

\subsubsection{Baselines}

We compare our method with the following two baselines:

\paragraph{Co-occurrences}

For every entity pair that occurs in at least one sentence in the dataset used to train the representation learning model, we predict the relation to be positive (\textit{i.e.} the concatenation of the types of the two entities). By design, this method will have perfect recall.

\paragraph{Attention} 

This method is similar to those presented by \citet{LSLLS2016} and \citet{BRAN2018}. For a given pair of entities, we collect every sentence containing the pair from the dataset used to train the representation learning model. Each sentence is passed to an LSTM whose final state is taken as the sentence representation. The representations for all sentences for the given entity pair are pooled together into a single representation using an attention mechanism. This representation is then used as the input to a feedforward network with a softmax function at the output. This method is therefore trained on exactly the same dataset as our pair-level classifier.

The attention model is trained for 1,000,000 iterations using a minibatch size of 100 (of which 50 are sampled from the \texttt{NO-RELATION} set). The parameters are optimised using Adam with a learning rate of 0.000005. As with our model, when making predictions on unseen data points, we only predict a label other than \texttt{NO-RELATION} if the predicted probability is higher than a threshold. This threshold is tuned to maximise the F1-score on the validation set.

\subsubsection{Results}

\begin{table}[t]
    \centering
    \begin{tabular}{lccc}
        \toprule
        \textsc{Model} & P & R & F \\
        \midrule
        Co-occurrences & 3.10 & 100.00 & 6.02 \\
        Attention & 11.06 & 26.97 & 15.69 \\
        Ours & 12.54 & 25.91 & 16.90 \\
        \bottomrule
    \end{tabular}
    \caption{Results on the test set of the pair-level classification task. We report the precision (P), recall (R) and the F1-score (F). For all metrics, higher is better.}
    \label{tab:experiments:pair:results}
    \vskip -0.1in
\end{table}

The precision, recall, and F1-score on the test set are reported in Table \ref{tab:experiments:pair:results}. Our model achieves a higher F1-score than the attention model. Unsurprisingly, both the attention model and our model achieve significantly higher precision than the co-occurrence baseline at the expense of lower recall. 

In contrast to the attention model, when classifying a new pair, our model does not need to encode all of the sentences containing that pair. This provides significant computational advantages, both in terms of memory and time.

\section{Conclusion}

We have presented a model for learning representations of pairs of biomedical entities from unlabelled text corpora. We use a latent variable with an arbitrarily flexible distribution in order to be able to capture the complex relations between each pair of entities. The unified architecture can be used for both mention-level and pair-level relation extraction. On both tasks, we achieve results competitive with strong baselines. We also show significant computational gains in terms of the number of parameters and training times.

Our model presents many avenues for future work. The results in Table \ref{tab:experiments:supervised:results} show that the model's performance improves with the size of the hidden states in the networks; this suggests that there are further gains achievable simply by providing the model with more parameters. The model could be further scaled up by using a hierarchy of latent variables to increase the expressive power of the representations. 

Other directions include evaluating the benefits of having a representation which explicitly captures uncertainty about the relations. For example, this can be done by assessing if the model is less confident when making predictions about entity pairs which do not occur frequently in the unlabelled corpus. Additionally, since our model can produce a representation for any pair of entities (even those which do not occur together in the unlabelled corpus), it could be used in a link prediction setting to score unseen entity pairs.

\section*{Acknowledgements}
We would like to thank our colleagues Sia Togia and Angus Brayne for their thorough feedback on this paper and Rogier Hintzen for his precious help in preparing the datasets.   

\bibliographystyle{acl_natbib}
\bibliography{refs}

\begin{thebibliography}{32}
\expandafter\ifx\csname natexlab\endcsname\relax\def\natexlab#1{#1}\fi

\bibitem[{Beltagy et~al.(2019)Beltagy, Lo, and Cohan}]{BLC2019}
I.~Beltagy, K.~Lo, and A.~Cohan. 2019.
\newblock {SciBERT: A Pretrained Language Model for Scientific Text}.
\newblock In \emph{Proceedings of the 2019 Conference on Empirical Methods in
  Natural Language Processing}.

\bibitem[{Bengio et~al.(2013)Bengio, Courville, and Vincent}]{BCV2013}
Y.~Bengio, A.~Courville, and P.~Vincent. 2013.
\newblock Representation learning: A review and new perspectives.
\newblock \emph{IEEE Transactions on Pattern Analysis and Machine
  Intelligence}, 35.

\bibitem[{Bowman et~al.(2016)Bowman, Vilnis, Vinyals, Dai, Jozefowicz, and
  Bengio}]{BVVDJB2016}
S.~Bowman, L.~Vilnis, O.~Vinyals, A.~Dai, R.~Jozefowicz, and S.~Bengio. 2016.
\newblock {Generating Sentences from a Continuous Space}.
\newblock In \emph{Proceedings of The 20th {SIGNLL} Conference on Computational
  Natural Language Learning}.

\bibitem[{Bravo et~al.(2015)Bravo, Pi{\~{n}}ero, Queralt-Rosinach, Rautschka,
  and Furlong}]{BPQRF2015}
{\`{A}}.~Bravo, J.~Pi{\~{n}}ero, N.~Queralt-Rosinach, M.~Rautschka, and
  L.~Furlong. 2015.
\newblock {Extraction of Relations Between Genes and Diseases from Text and
  Large-Scale Data Analysis: Implications for Translational Research}.
\newblock \emph{BMC Bioinformatics}, 16.

\bibitem[{Camacho-Collados et~al.(2019)Camacho-Collados, Espinosa-Anke, Jameel,
  and Schockaert}]{ijcai2019relationvectors}
J.~Camacho-Collados, L.~Espinosa-Anke, S.~Jameel, and S.~Schockaert. 2019.
\newblock {A Latent Variable Model for Learning Distributional Relation
  Vectors}.
\newblock In \emph{Proceedings of the Twenty-Eighth International Joint
  Conference on Artificial Intelligence}.

\bibitem[{Devlin et~al.(2019)Devlin, Chang, Lee, and Toutanova}]{DCLT2019}
J.~Devlin, M.~Chang, K.~Lee, and K.~Toutanova. 2019.
\newblock {BERT: Pre-training of Deep Bidirectional Transformers for Language
  Understanding}.
\newblock In \emph{Proceedings of the 2019 Conference of the North American
  Chapter of the Association for Computational Linguistics}.

\bibitem[{Graves(2013)}]{Graves_RNN_2013}
A.~Graves. 2013.
\newblock {Generating Sequences With Recurrent Neural Networks}.
\newblock \emph{CoRR}, abs/1308.0850.

\bibitem[{Hoffmann et~al.(2011)Hoffmann, Zhang, Ling, Zettlemoyer, and
  Weld}]{multir2011}
R.~Hoffmann, C.~Zhang, X.~Ling, L.~Zettlemoyer, and D.~Weld. 2011.
\newblock {Knowledge-Based Weak Supervision for Information Extraction of
  Overlapping Relations}.
\newblock In \emph{Proceedings of the 49th Annual Meeting of the Association
  for Computational Linguistics}.

\bibitem[{Jia et~al.(2019)Jia, Wong, and Poon}]{JWP2019}
R.~Jia, C.~Wong, and H.~Poon. 2019.
\newblock {Document-Level N-ary Relation Extraction with Multiscale
  Representation Learning}.
\newblock In \emph{Proceedings of the 2019 Conference of the North American
  Chapter of the Association for Computational Linguistics}.

\bibitem[{Joshi et~al.(2019)Joshi, Choi, Levy, Weld, and
  Zettlemoyer}]{pair2vec2019}
M.~Joshi, E.~Choi, O.~Levy, D.~Weld, and L.~Zettlemoyer. 2019.
\newblock {pair2vec: Compositional Word-Pair Embeddings for Cross-Sentence
  Inference}.
\newblock In \emph{Proceedings of the 2019 Conference of the North American
  Chapter of the Association for Computational Linguistics}.

\bibitem[{Kingma and Ba(2015)}]{KB2015}
D.~Kingma and J.~Ba. 2015.
\newblock {Adam: A Method for Stochastic Optimization}.
\newblock In \emph{International Conference on Learning Representations}.

\bibitem[{Kingma and Welling(2014)}]{KW2014}
D.~Kingma and M.~Welling. 2014.
\newblock {Auto-Encoding Variational Bayes}.
\newblock In \emph{International Conference on Learning Representations}.

\bibitem[{Krallinger et~al.(2017)Krallinger, Rabal, Akhondi, P{\'e}rez,
  Santamar{\'i}a, Rodr{\'i}guez, Tsatsaronis, Intxaurrondo, L{\'o}pez, Nandal,
  van Buel, Chandrasekhar, Rodenburg, L{\ae}greid, Doornenbal, Oyarz{\'a}bal,
  Lourenço, and Valencia}]{KRAPSRTILNVCRLDOLV2017}
M.~Krallinger, O.~Rabal, S.~Akhondi, M.~P{\'e}rez, J.~Santamar{\'i}a,
  G.~Rodr{\'i}guez, G.~Tsatsaronis, A.~Intxaurrondo, J.~L{\'o}pez, U.~Nandal,
  E.~van Buel, A.~Chandrasekhar, M.~Rodenburg, A.~L{\ae}greid, M.~Doornenbal,
  J.~Oyarz{\'a}bal, A.~Lourenço, and A.~Valencia. 2017.
\newblock {Overview of the BioCreative VI chemical-protein interaction Track}.
\newblock In \emph{Proceedings of the BioCreative VI Workshop}.

\bibitem[{Lee et~al.(2019)Lee, Yoon, Kim, Kim, Kim, So, and Kang}]{LYKKKSK2019}
J.~Lee, W.~Yoon, S.~Kim, D.~Kim, S.~Kim, C.~So, and J.~Kang. 2019.
\newblock {BioBERT: A Pre-Trained Biomedical Language Representation Model for
  Biomedical Text Mining}.
\newblock \emph{Bioinformatics}.

\bibitem[{Liang et~al.(2019)Liang, Li, Song, Madden, Ding, and
  Bu}]{knowledge-graph-cascade-2019}
X.~Liang, D.~Li, M.~Song, A.~Madden, Y.~Ding, and Y.~Bu. 2019.
\newblock {Predicting Biomedical Relationships Using the Knowledge and Graph
  Embedding Cascade Model}.
\newblock \emph{PLOS ONE}, 14.

\bibitem[{Lin et~al.(2016)Lin, Shen, Liu, Luan, and Sun}]{LSLLS2016}
Y.~Lin, S.~Shen, Z.~Liu, H.~Luan, and M.~Sun. 2016.
\newblock {Neural Relation Extraction with Selective Attention over Instances}.
\newblock In \emph{Proceedings of the 54th Annual Meeting of the Association
  for Computational Linguistics}.

\bibitem[{Mattei and Frellsen(2018)}]{MF2018}
P.~Mattei and J.~Frellsen. 2018.
\newblock {Leveraging the Exact Likelihood of Deep Latent Variable Models}.
\newblock In \emph{Advances in Neural Information Processing Systems 31}.

\bibitem[{Melis et~al.(2018)Melis, Dyer, and Blunsom}]{SOTALM_Melis_2018}
G.~Melis, C.~Dyer, and P.~Blunsom. 2018.
\newblock {On the State of the Art of Evaluation in Neural Language Models}.
\newblock In \emph{International Conference on Learning Representations}.

\bibitem[{Mintz et~al.(2009)Mintz, Bills, Snow, and
  Jurafsky}]{mintz-etal-2009-distant}
M.~Mintz, S.~Bills, R.~Snow, and D.~Jurafsky. 2009.
\newblock {Distant Supervision for Relation Extraction Without Labeled Data}.
\newblock In \emph{Proceedings of the Joint Conference of the 47th Annual
  Meeting of the ACL and the 4th International Joint Conference on Natural
  Language Processing of the AFNLP}.

\bibitem[{van Mulligen et~al.(2012)van Mulligen, Fourrier-Reglat, Gurwitz,
  Molokhia, Nieto, Trifiro, Kors, and Furlong}]{VFGMNTKF2012}
E.~van Mulligen, A.~Fourrier-Reglat, D.~Gurwitz, M.~Molokhia, A.~Nieto,
  G.~Trifiro, J.~Kors, and L.~Furlong. 2012.
\newblock {The EU-ADR Corpus}.
\newblock \emph{Journal of Biomedical Informatics}, 45.

\bibitem[{Percha and Altman(2015)}]{ebc-altman-2015}
B.~Percha and R.~Altman. 2015.
\newblock {Learning the Structure of Biomedical Relationships from Unstructured
  Text}.
\newblock \emph{PLoS Computational Biology}, 11.

\bibitem[{Peters et~al.(2018)Peters, Neumann, Iyyer, Gardner, Clark, Lee, and
  Zettlemoyer}]{peters-etal-2018-deep}
M.~Peters, M.~Neumann, M.~Iyyer, M.~Gardner, C.~Clark, K.~Lee, and
  L.~Zettlemoyer. 2018.
\newblock {Deep Contextualized Word Representations}.
\newblock In \emph{Proceedings of the 2018 Conference of the North {A}merican
  Chapter of the Association for Computational Linguistics}.

\bibitem[{Quirk and Poon(2017)}]{QuirkPoon16}
C.~Quirk and H.~Poon. 2017.
\newblock {Distant Supervision for Relation Extraction beyond the Sentence
  Boundary}.
\newblock In \emph{Proceedings of the 15th Conference of the European Chapter
  of the Association for Computational Linguistics}.

\bibitem[{Rezende et~al.(2014)Rezende, Mohamed, and Wierstra}]{RMW2014}
D.~Rezende, S.~Mohamed, and D.~Wierstra. 2014.
\newblock {Stochastic Backpropagation and Approximate Inference in Deep
  Generative Models}.
\newblock In \emph{Proceedings of the 31st International Conference on Machine
  Learning}.

\bibitem[{Segura-Bedmar et~al.(2013)Segura-Bedmar, Mart{\'\i}nez, and
  Herrero-Zazo}]{SMH2013}
I.~Segura-Bedmar, P.~Mart{\'\i}nez, and M.~Herrero-Zazo. 2013.
\newblock {SemEval-2013 Task 9: Extraction of Drug-Drug Interactions from
  Biomedical Texts (DDIExtraction 2013)}.
\newblock In \emph{Second Joint Conference on Lexical and Computational
  Semantics (*{SEM}), Volume 2: Proceedings of the Seventh International
  Workshop on Semantic Evaluation}.

\bibitem[{Soares et~al.(2019)Soares, FitzGerald, Ling, and
  Kwiatkowski}]{SFLK2019}
L.~Soares, N.~FitzGerald, J.~Ling, and T.~Kwiatkowski. 2019.
\newblock {Matching the Blanks: Distributional Similarity for Relation
  Learning}.
\newblock In \emph{Proceedings of the 57th Annual Meeting of the Association
  for Computational Linguistics}.

\bibitem[{Vaswani et~al.(2017)Vaswani, Shazeer, Parmar, Uszkoreit, Jones,
  Gomez, Kaiser, and Polosukhin}]{VSPUJGKP2017}
A.~Vaswani, N.~Shazeer, N.~Parmar, J.~Uszkoreit, L.~Jones, A.~Gomez,
  {\L}~Kaiser, and I.~Polosukhin. 2017.
\newblock {Attention is All you Need}.
\newblock In \emph{Advances in Neural Information Processing Systems}.

\bibitem[{Verga et~al.(2018)Verga, Strubell, and McCallum}]{BRAN2018}
P.~Verga, E.~Strubell, and A.~McCallum. 2018.
\newblock {Simultaneously Self-Attending to All Mentions for Full-Abstract
  Biological Relation Extraction}.
\newblock In \emph{North American Chapter of the Association for Computational
  Linguistics}.

\bibitem[{Zhang et~al.(2018)Zhang, Lin, Yang, Wang, Zhang, Sun, and
  Yang}]{zhang-hybrid-2018}
Y.~Zhang, H.~Lin, Z.~Yang, J.~Wang, S.~Zhang, Y.~Sun, and L.~Yang. 2018.
\newblock {A Hybrid Model Based on Neural Networks for Biomedical Relation
  Extraction}.
\newblock \emph{Journal of Biomedical Informatics}, 81.

\bibitem[{Zhang et~al.(2017)Zhang, Zhong, Chen, Angeli, and
  Manning}]{tacred2017}
Y.~Zhang, V.~Zhong, D.~Chen, G.~Angeli, and C.~Manning. 2017.
\newblock {Position-aware Attention and Supervised Data Improve Slot Filling}.
\newblock In \emph{Proceedings of the 2017 Conference on Empirical Methods in
  Natural Language Processing}.

\bibitem[{Zhang et~al.(2020)Zhang, Shu, Yu, Liu, Zhao, Li, and
  Guo}]{Distill_Zhang_2020}
Zhenyu Zhang, Xiaobo Shu, Bowen Yu, Tingwen Liu, Jiapeng Zhao, Quangang Li, and
  Li~Guo. 2020.
\newblock {Distilling Knowledge from Well-Informed Soft Labels for Neural
  Relation Extraction}.
\newblock In \emph{The Thirty-Fourth AAAI Conference on Artificial
  Intelligence}.

\bibitem[{Zhou et~al.(2016)Zhou, Qi, Zheng, Xu, Bao, and
  Xu}]{zhou-etal-2016-text}
P.~Zhou, Z.~Qi, S.~Zheng, J.~Xu, H.~Bao, and B.~Xu. 2016.
\newblock {Text Classification Improved by Integrating Bidirectional LSTM with
  Two-dimensional Max Pooling}.
\newblock In \emph{Proceedings of the 26th International Conference on
  Computational Linguistics}.

\end{thebibliography}

\end{document}